\def\year{2019}\relax
\documentclass[letterpaper]{article} %DO NOT CHANGE THIS
\usepackage{aaai19}  %Required
\usepackage{times}  %Required
\usepackage{helvet}  %Required
\usepackage{courier}  %Required
\usepackage{url}  %Required
\usepackage{graphicx}  %Required

\usepackage{epsfig}
\usepackage{amsmath}
\usepackage{amssymb}
\usepackage{color}
\usepackage{comment}
\usepackage{algorithm}
\usepackage{algorithmic}

\usepackage{enumitem}
\usepackage{amsfonts}
\usepackage{extarrows}
\usepackage{multirow}
\usepackage{rotating}

\begin{document}
\title{Unsupervised Learning of Neural Networks to Explain Neural Networks (extended abstract)}
\author{Quanshi Zhang$^{a}$, Yu Yang$^{b}$, and Ying Nian Wu$^{b}$\\
$^{a}$Shanghai Jiao Tong University,\qquad$^{b}$University of California, Los Angeles}
\maketitle

\section{Introduction}

\begin{figure}[t]
\centering
\includegraphics[width=0.95\linewidth]{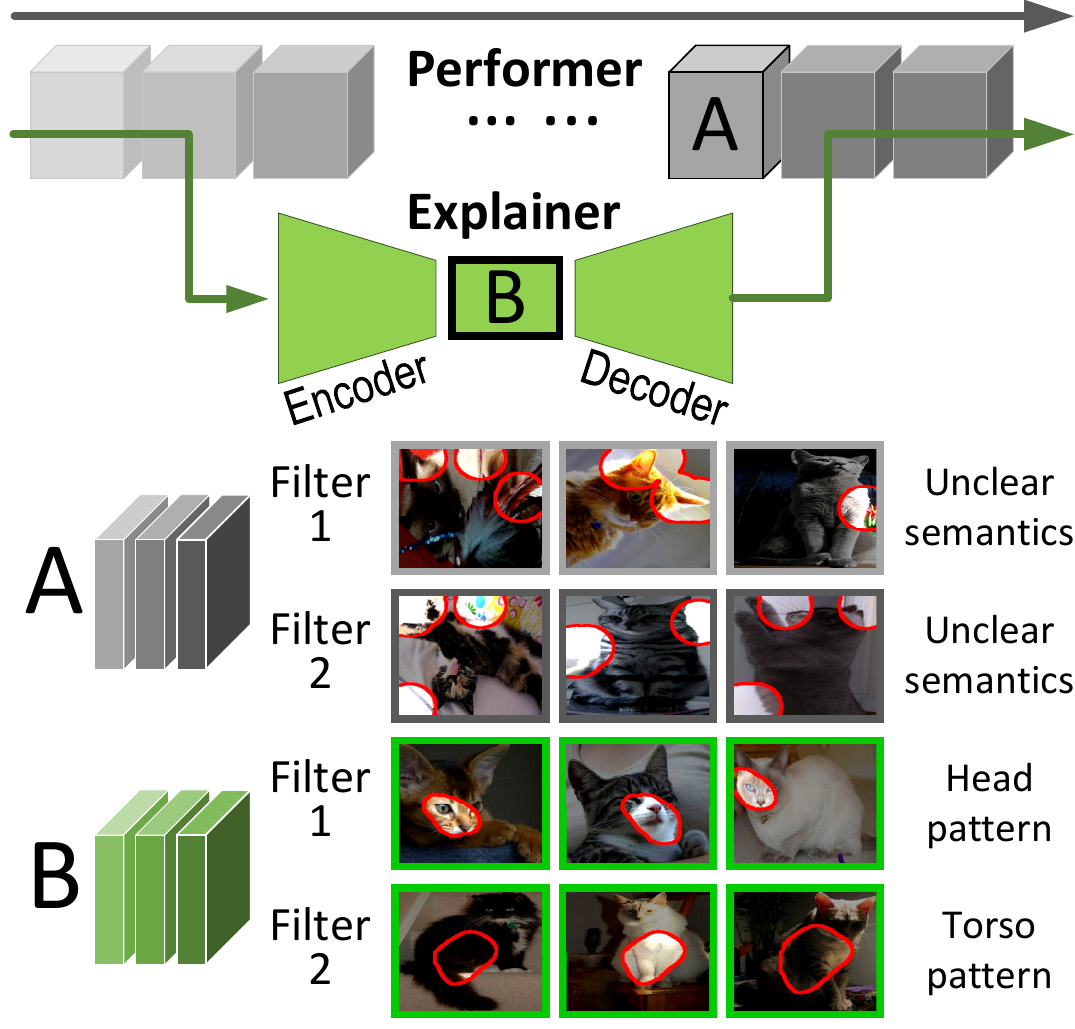}
\vspace{0pt}
\caption{Explainer network. We use an explainer network (green) to disentangle the feature map of a certain conv-layer in a pre-trained performer network (gray). The explainer network disentangles input features into object-part feature maps ($B$) to explain knowledge representations in the performer, \emph{i.e.} making each filter represent a specific object part. The explainer network can also invert the disentangled object-part features to reconstruct features of the performer without much loss of information. We compare ordinary feature maps ($A$) in the performer and the disentangled feature maps ($B$) in the explainer on the right. The gray and green lines indicate the information-pass route during the inference process and that during the explanation process, respectively.}
\label{fig:top}
\vspace{-10pt}
\end{figure}

\footnote[1]{Quanshi Zhang is with the John Hopcroft Center and the MoE Key Lab of Artificial Intelligence, AI Institute, Shanghai Jiao Tong University. Yu Yang and Ying Nian Wu are with the Center for Vision, Cognition, Learning, and Autonomy, University of California, Los Angeles.}\textbf{Motivation, diagnosis of features inside CNNs:} In recent years, real applications usually propose new demands for deep learning beyond the accuracy. The CNN needs to earn trust from people for safety issues, because a high accuracy on testing images cannot always ensure that the CNN encodes correct features. Instead, the CNN sometimes uses unreliable reasons for prediction.

Therefore, this study aim to provide a generic tool to examine middle-layer features of a CNN to ensure the safety in critical applications. Unlike previous visualization~\cite{CNNVisualization_1} and diagnosis~\cite{Interpretability,trust} of CNN representations, we focus on the following two new issues, which are of special values in feature diagnosis.

\noindent
$\bullet\;$Disentanglement of interpretable and uninterpretable feature information is necessary for a rigorous and trustworthy examination of CNN features. Each filter of a conv-layer usually encodes a mixture of various semantics and noises (see Fig.~\ref{fig:top}). As discussed in \cite{Interpretability}, filters in high conv-layers mainly represent ``object parts''\footnote[2]{\cite{interpretableCNN} considers both semantics of ``objects'' and ``parts'' as parts.}, and ``material'' and ``color'' information in high layers is not salient enough for trustworthy analysis. In particular, part features are usually more localized and thus is more helpful in feature diagnosis.

Therefore, in this paper, we propose to disentangle part features from another signals and noises. For example, we may quantitatively disentangle 90\% information of CNN features as object parts and interpret the rest 10\% as textures and noises.

\begin{figure}[t]
\centering
\includegraphics[width=0.99\linewidth]{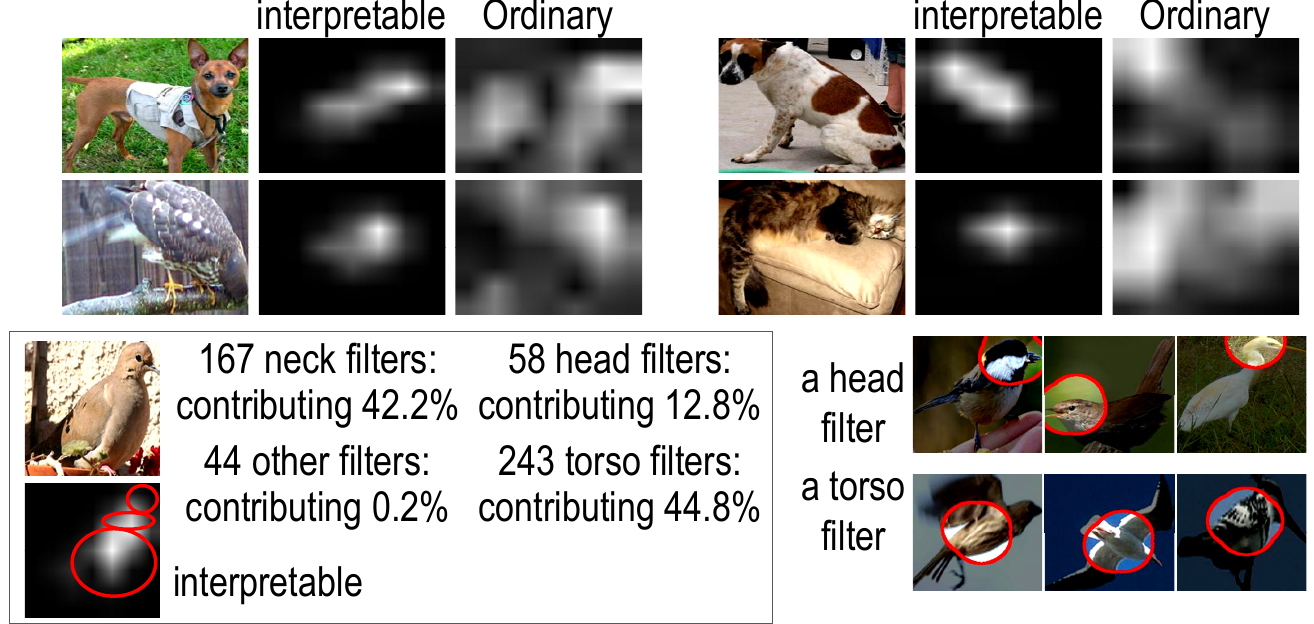}
\vspace{0pt}
\caption{Grad-CAM attention maps and quantitative analysis. We compute grad-CAM attention maps of interpretable feature maps in the explainer and ordinary feature maps in the performer.}
\label{fig:heatmap_analysis}
\vspace{-10pt}
\end{figure}

\noindent
$\bullet\;$Semantic explanations: Given an input image, we aim to use clear visual concepts (here, object parts) to interpret chaotic CNN features. In comparisons, network visualization and diagnosis mainly illustrate the appearance corresponding to a network output/filter, without physically modeling or quantitatively summarizing strict semantics. As shown in Fig.~\ref{fig:heatmap_analysis}, our method identifies which parts are learned and used for the prediction as more fine-grained explanations for CNN features.

\begin{table}
\begin{center}
\renewcommand\tabcolsep{3pt}
\resizebox{\linewidth}{!}{\begin{tabular}{lp{1.8cm}p{1.7cm}p{2.3cm}p{2cm}}
\!\!\!&\!\!\! Disentangle interpretable signals \!\!\!&\!\!\! Semantically explain \!\!\!&\!\!\! Few restrictions on structures \& losses \!\!\!&\!\!\! Not affect discriminability\!\!\!\\
\hline
\!\!\!CNN visualization\!\!\! & & & $\qquad\checkmark$ & $\qquad\checkmark$\\
\!\!\!Interpretable nets\!\!\! & $\qquad\checkmark$ & $\qquad\checkmark$ & &\\
\!\!\!Our research & $\qquad\checkmark$ & $\qquad\checkmark$ & $\qquad\checkmark$ & $\qquad\checkmark$\\
\hline
\end{tabular}}
\vspace{2pt}
\caption{Comparison between our research and other studies. Note that this table can only summarize mainstreams in different research directions considering the huge research diversity.\vspace{-20pt}}
\label{tab:diff}
\end{center}
\end{table}

\textbf{Tasks, learning networks to explain networks:} In this paper, we propose a new explanation strategy to boost feature interpretability. \emph{I.e.} given a pre-trained CNN, we learn another neural network, namely a \textit{explainer} network, to translate chaotic middle-layer features of the CNN into semantically meaningful object parts. More specifically, as shown in Fig.~\ref{fig:top}, the explainer decomposes middle-layer feature maps into elementary feature components of object parts. Accordingly, the pre-trained CNN is termed a \textit{performer} network.

In the scenario of this study, the performer is well pre-trained for superior performance. We attach the explainer onto the performer without affecting the original discrimination power of the performer.

The explainer works like an auto-encoder. The encoder decomposes features in the performer into interpretable part features and other uninterpretable features. The encoder contains hundreds of specific filters, each being learned to represent features of a certain object part. The decoder inverts the disentangled part features to reconstruct features of upper layers of the performer.

As shown in Fig.~\ref{fig:top}, the feature map of each filter in the performer usually represents a chaotic mixture of object parts and textures, whereas the disentangled object-part features in the explainer can be treated as a paraphrase of performer features that provide an insightful understanding of the performer. For example, the explainer can tell us\\
$\bullet\;$How much can feature information (\emph{e.g.} 90\%) in the performer be interpreted as object parts?\\
$\bullet\;$Information of what parts is encoded in the performer?\\
$\bullet\;$For each specific prediction, which object parts activate filters in the performer, and how much do they contribute to the prediction?

\textbf{Explaining black-box networks vs. learning interpretable networks:} In recent years, some researchers gradually focus on the interpretability~\cite{Interpretability} of middle-layer features of a neural network. Pioneering studies, such as the research of capsule nets~\cite{capsule} and interpretable CNNs~\cite{interpretableCNN}, have developed new algorithms to ensure middle-layer features of a neural network are semantic meaningful.

In comparisons, our explaining pre-trained performer is of higher flexibility and has broader applicability than learning new interpretable models. Table~\ref{tab:diff} summarizes the difference.\\
$\bullet\;$Model flexibility: Interpretable neural networks usually have specific requirements for structures~\cite{capsule} or losses~\cite{interpretableCNN}, which limit the model flexibility and applicability. In addition, most existing CNNs are learned in a black-box manner with low interpretability. To interpret such CNNs, an explainer is required.\\
$\bullet\;$Interpretability vs. discriminability: Using clear visual concepts to explain black-box networks can overcome a major issue with network interpretability, \emph{i.e.} the dilemma between the feature interpretability and its discrimination power. A high interpretability is not necessarily equivalent to, and sometimes conflicts with a high discrimination power~\cite{Interpretability}. As discussed in \cite{interpretableCNN}, increasing the interpretability of a neural network may affect its discrimination power. People usually have to trade off between the network interpretability and the performance in real applications.

In contrast, our explanation strategy does not change feature representations in the pre-trained CNN performer, thereby physically protecting the CNN's discrimination power.

\begin{figure*}[t]
\centering
\includegraphics[width=0.9\linewidth]{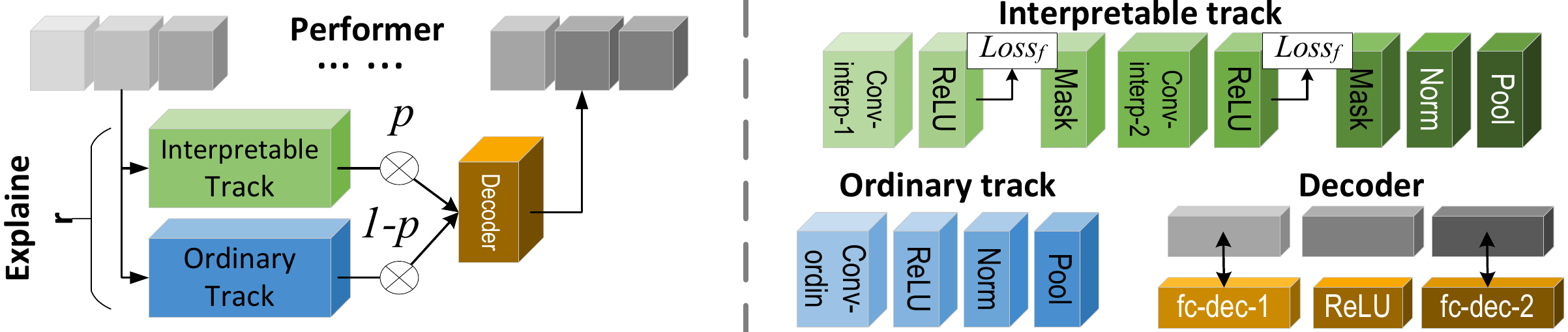}
\vspace{0pt}
\caption{The explainer network (left). Detailed structures within the interpretable track, the ordinary track, and the decoder are shown on the right. People can change the number of conv-layers and FC layers within the encoder and the decoder for their own applications.}
\label{fig:block}
\vspace{-10pt}
\end{figure*}

\textbf{Learning:} We learn the explainer by distilling feature representations from the performer to the explainer without any additional supervision. No annotations of parts or textures are used to guide the feature disentanglement during the learning process. We add a loss to specific filters in the explainer (see Fig.~\ref{fig:block}). The filter loss encourages the filter to be exclusively triggered by a certain object part of a category. This filter is termed an interpretable filter.

Meanwhile, the disentangled object-part features are also required to reconstruct features of upper layers of the performer. Successful feature reconstructions guarantee to avoid significant information loss during the disentanglement of part features.

\textbf{Contributions} of this study are summarized as follows.

\noindent
(i) We tackle a new explanation strategy, \emph{i.e.} learning an explainer network to mine and clarify potential feature components in middle layers of a pre-trained performer network. Decomposing chaotic middle-layer features into interpretable concepts will shed new light on explaining black-box models.

\noindent
(ii) Another distinctive contribution of this study is that learning an explainer for interpretation avoids the typical dilemma between a model's discriminability and interpretability. This is our essential difference to studies of directly learning the performer with disentangled/interpretable features.

Our method protects the discrimination power of the original network. Thus, it ensures a high flexibility and broad applicability in real applications.

\noindent
(iii) Our method is able to learn the explainer without any annotations of object parts or textures for supervision. Experiments show that our approach has considerably improved the feature interpretability.

\section{Algorithm}

\subsection{Network structure of the explainer}

As shown in Fig.~\ref{fig:block}, the explainer network has two modules, \emph{i.e.} an encoder and a decoder, which transform performer features into interpretable object-part features and invert object-part features back to features of the performer, respectively. We can roughly consider that object-part features in the explainer contain nearly the same information as features in the performer.

We applied the encoder and decoder with following structures to all types of performers in all experiments. Nevertheless, people can change the layer number of the explainer in their applications.

\textbf{Encoder:} In order to reduce the risk of over-interpreting textures or noises as parts, we design two tracks for the encoder, namely an \textit{interpretable track} and an \textit{ordinary track}, which models part features and other features, respectively. Although as discussed in \cite{interpretableCNN}, a high conv-layer mainly represents parts rather than textures, avoiding over-interpreting is still necessary for the explainer.

The interpretable track disentangles performer features into object parts. This track has two interpretable conv-layers (namely \textit{conv-interp-1,conv-interp-2}), each followed by a ReLU layer and a mask layer. The interpretable layer contains interpretable filters. Each interpretable filter is learned based on the filter loss, which makes the filter exclusively triggered by a specific object part (the learning of interpretable filters will be introduced later). The ordinary track contains a conv-layer (namely \textit{conv-ordin}), a ReLU layer, and a pooling layer.

We sum up output features of the interpretable track {\small$x_{\textrm{interp}}$} and those of the ordinary track {\small$x_{\textrm{ordin}}$} as the final output of the encoder, \emph{i.e.} {\small$x_{\textrm{enc}}=p\cdot x_{\textrm{interp}}+(1-p)\cdot x_{\textrm{ordin}}$}, where a scalar weight $p$ measures the quantitative contribution from the interpretable track. $p$ is parameterized as a softmax probability {\small$p=sigmoid(w_{p})$}, $w_{p}\in{\boldsymbol\theta}$, where ${\boldsymbol\theta}$ is the set of parameters to be learned. Our method encourages a large $p$ so that most information in {\small$x_{\textrm{enc}}$} comes from the interpretable track.

In particular, if $p=0.9$, we can roughly consider that about 90\% feature information from the performer can be represented as object parts due to the use of norm-layers.

\textbf{Decoder:} The decoder inverts {\small$x_{\textrm{enc}}$} to {\small$x_{\textrm{dec}}$}, which reconstructs performer features. The decoder has two FC layers, which followed by two ReLU layers. We use the two FC layers, namely \textit{fc-dec-1} and \textit{fc-dec-2}, to reconstruct feature maps of two corresponding FC layers in the performer. The reconstruction loss will be introduced later. The better reconstruction of the FC features indicates that the explainer loses less information during the computation of {\small$x_{\textrm{enc}}$}.

\subsection{Learning}

When we distill knowledge representations from the performer to the explainer, we consider the following three terms: 1) the quality of knowledge distillation, \emph{i.e.} the explainer needs to well reconstruct feature maps of upper layers in the performer, thereby minimizing the information loss; 2) the interpretability of feature maps of the interpretable track, \emph{i.e.} each filter in \textit{conv-interp-2} should exclusively represent a certain object part; 3) the relative contribution of the interpretable track \emph{w.r.t.} the ordinary track, \emph{i.e.} we hope the interpretable track to make much more contribution to the final CNN prediction than the ordinary track. Therefore, we minimize the following loss for each input image to learn the explainer.
\begin{small}
\vspace{-2pt}
\begin{equation}
\!Loss({\boldsymbol\theta})\!=\!\sum_{l\in{\bf L}}\lambda_{(l)}\!\Vert x_{(l)}\!-\!x_{(l)}^{*}\Vert^2\!\!-\!\eta\log p\!+\!\!\sum_{f}\lambda_{f}\!\cdot\!Loss_{f}(x_{f})\!\!
\label{eqn:loss}
\vspace{-2pt}
\end{equation}
\end{small}
where ${\boldsymbol\theta}$ denotes the set of parameters to be learned, including filter weights of conv-layers and FC layers in the explainer, $w_{p}$ for $p$, and ${\boldsymbol\alpha}$ for norm-layers. $\lambda_{(l)},\lambda_{f}$ and $\eta$ are scalar weights.

\textbf{The first term} {\small$\Vert x_{(l)}-x_{(l)}^{*}\Vert^2$} is the reconstruction loss, where $x_{(l)}$ denotes the feature of the FC layer $l$ in the decoder, {\small${\bf L}=\{fc-dec-1,fc-dec-2\}$}. {\small$x_{(l)}^{*}$} indicates the corresponding feature in the performer.

\textbf{The second term} $-\log p$ encourages the interpretable track to make more contribution to the CNN prediction.

\textbf{The third term} {\small$Loss_{f}(x_{f})$} is the loss of filter interpretability. Without annotations of object parts, the filter loss forces $x_{f}$ to be exclusively triggered by a specific object part of a certain category. The filter loss was formulated in \cite{interpretableCNN}. We can summarize the filter loss as the minus mutual information between the distribution of feature maps and that of potential part locations.

\section{Experiments}

In experiments, we learned explainers for performer networks with three types of structures to demonstrate the broad applicability of our method. Performer networks were pre-trained using object images in two different benchmark datasets for object classification. We visualized feature maps of interpretable filters in the explainer to illustrate semantic meanings of these filters. Experiments showed that interpretable filters in the explainer generated more semantically meaningful feature maps than conv-layers in the performer.

\begin{figure}[t]
\centering
\includegraphics[width=0.99\linewidth]{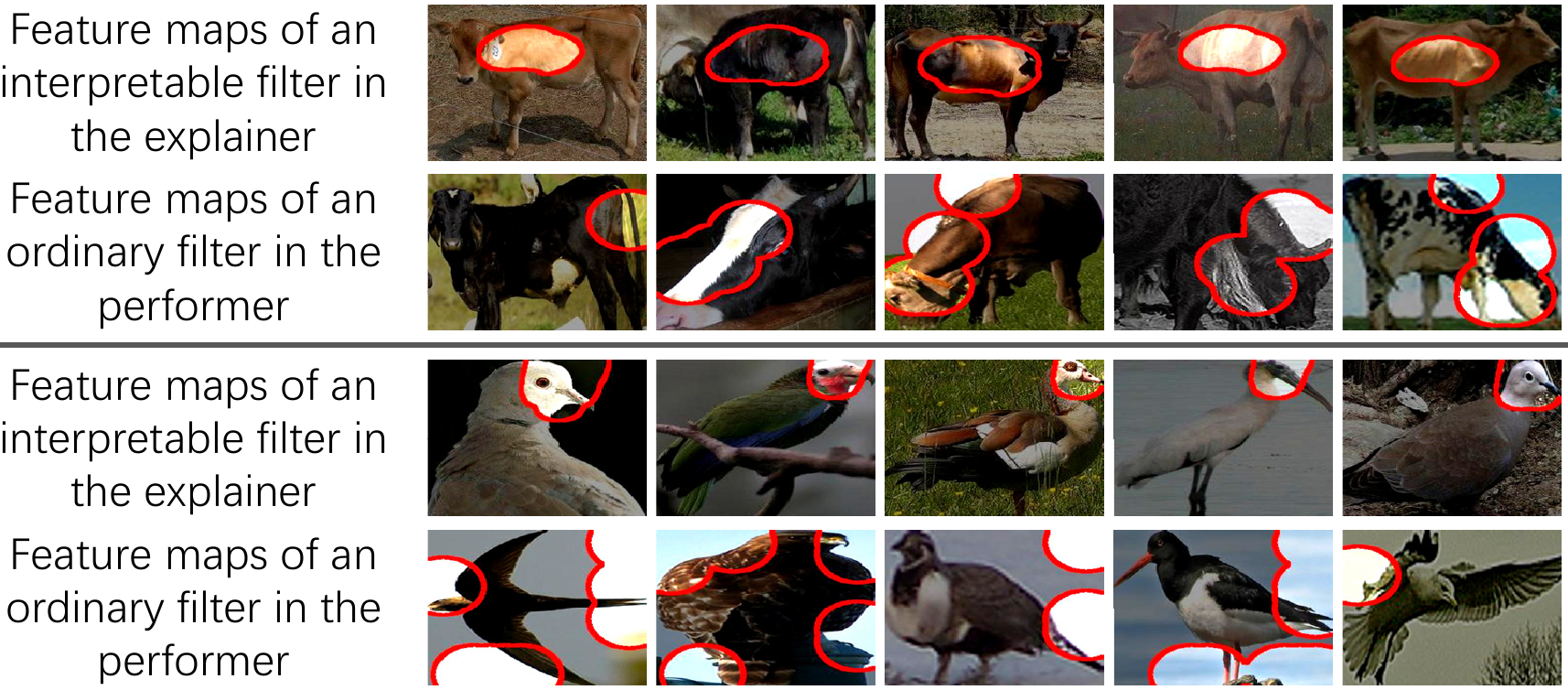}
\vspace{0pt}
\caption{Visualization of interpretable filters in the explainer and ordinary filters in the performer. We compared filters in the top conv-layer of the performer and interpretable filters in the \textit{conv-interp-2} layer of the explainer.}
\label{fig:visual}
\vspace{-10pt}
\end{figure}

We compared the object-part interpretability between feature maps of the explainer and those of the performer. To obtain a convincing evaluation, we both visualized filters (see Fig.~\ref{fig:visual}) and used the objective metric of location instability~\cite{interpretableCNN} to measure the fitness between a filter $f$ and the representation of an object part.

\begin{table}
\begin{center}
\resizebox{\linewidth}{!}{\begin{tabular}{l|ccccccc|c}
\hline
&\multicolumn{7}{|c|}{Single-category} & Multi\\
\hline
\!\!&\!\! bird \!\!&\!\! cat \!\!&\!\! cow \!\!&\!\! dog \!\!&\!\! horse \!\!&\!\! sheep \!\!&\!\! Avg. \!\!&\!\! Avg.\\
\hline
AlexNet \!\!&\!\!0.153
\!\!&\!\!0.131
\!\!&\!\!0.141
\!\!&\!\!0.128
\!\!&\!\!0.145
\!\!&\!\!0.140
\!\!&\!\!\textcolor{blue}{0.140}
\!\!&\!\! --\\
Explainer \!\!&\!\!{\bf0.104}
\!\!&\!\!{\bf0.089}
\!\!&\!\!{\bf0.101}
\!\!&\!\!{\bf0.083}
\!\!&\!\!{\bf0.098}
\!\!&\!\!{\bf0.103}
\!\!&\!\!\textcolor{blue}{\bf0.096}
\!\!&\!\! --\\
\hline
VGG-M \!\!&\!\!0.152
\!\!&\!\!0.132
\!\!&\!\!0.143
\!\!&\!\!0.130
\!\!&\!\!0.145
\!\!&\!\!0.141
\!\!&\!\!\textcolor{blue}{0.141}
\!\!&\!\!0.135\\
Explainer \!\!&\!\!{\bf0.106}
\!\!&\!\!{\bf0.088}
\!\!&\!\!{\bf0.101}
\!\!&\!\!{\bf0.088}
\!\!&\!\!{\bf0.097}
\!\!&\!\!{\bf0.101}
\!\!&\!\!\textcolor{blue}{\bf0.097}
\!\!&\!\!{\bf0.097}\\
\hline
VGG-S \!\!&\!\!0.152
\!\!&\!\!0.131
\!\!&\!\!0.141
\!\!&\!\!0.128
\!\!&\!\!0.144
\!\!&\!\!0.141
\!\!&\!\!\textcolor{blue}{0.139}
\!\!&\!\!0.138\\
Explainer \!\!&\!\!{\bf0.110}
\!\!&\!\!{\bf0.085}
\!\!&\!\!{\bf0.098}
\!\!&\!\!{\bf0.085}
\!\!&\!\!{\bf0.091}
\!\!&\!\!{\bf0.096}
\!\!&\!\!\textcolor{blue}{\bf0.094}
\!\!&\!\!{\bf0.107}\\
\hline
VGG-16 \!\!&\!\!0.145
\!\!&\!\!0.133
\!\!&\!\!0.146
\!\!&\!\!0.127
\!\!&\!\!0.143
\!\!&\!\!0.143
\!\!&\!\!\textcolor{blue}{0.139}
\!\!&\!\! 0.128\\
Explainer \!\!&\!\!{\bf0.095}
\!\!&\!\!{\bf0.089}
\!\!&\!\!{\bf0.097}
\!\!&\!\!{\bf0.085}
\!\!&\!\!{\bf0.087}
\!\!&\!\!{\bf0.089}
\!\!&\!\!\textcolor{blue}{\bf0.090}
\!\!&\!\!{\bf0.109}\\
\hline
\end{tabular}}
\end{center}
\vspace{0pt}
\caption{Location instability of feature maps between performers and explainers that were trained using the Pascal-Part dataset. A low location instability indicates a high filter interpretability.}
\label{tab:instability_VOC}
\vspace{-10pt}
\end{table}

\begin{table}
\begin{center}
\resizebox{0.9\linewidth}{!}{\begin{tabular}{c|c|c|c|c}
\hline
& AlexNet & VGG-M & VGG-S & VGG-16\\
\hline
Performer & 0.1502 & 0.1476 & 0.1481 & 0.1373\\
Explainer & {\bf0.0906} & {\bf0.0815} & {\bf0.0704} & {\bf0.0490}\\
\hline
\end{tabular}}
\end{center}
\vspace{0pt}
\caption{Location instability of feature maps in performers and explainers that were trained using the CUB200-2011 dataset. A low location instability indicates a high filter interpretability.}
\label{tab:instability_CUB}
\vspace{-10pt}
\end{table}

Tables~\ref{tab:instability_VOC} and \ref{tab:instability_CUB} compare the interpretability between feature maps in the performer and feature maps in the explainer. Feature maps in our explainers were much more interpretable than feature maps in performers in all comparisons.

Table~\ref{tab:p} lists $p$ values of explainers that were learned for different performers. $p$ measures the quantitative contribution from the interpretable track. For example, the VGG-16 network learned using the CUB200-2011 dataset has a $p$ value $p=0.9579$, which means that about $96\%$ feature information of the performer can be represented as object parts, and only about $4\%$ feature information comes from textures and noises.

\begin{table}
\begin{center}
\resizebox{0.9\linewidth}{!}{\begin{tabular}{l|cc|c}
\hline
& \multicolumn{2}{c|}{{\small Pascal-Part dataset}} &\!\!\! {\small CUB200}\!\!\\
\cline{2-3}
& {\footnotesize Single-class} \!\!\!&\!\!\! {\footnotesize Multi-class} \!\!\!&\!\!\! {\small dataset}\!\!\\
\hline
\!\!{\small AlexNet}
&--
&0.7137
&0.5810\\
\!\!{\small VGG-M}
&0.9012
&0.8066
&0.8611\\
\!\!{\small VGG-S}
&0.9270
&0.8996
&0.9533\\
\!\!{\small VGG-16}
&0.8593
&0.8718
&0.9579\\
\hline
\end{tabular}}
\vspace{0pt}
\end{center}
\caption{Average $p$ values of explainers. $p$ measures the quantitative contribution from the interpretable track. When we used an explainer to interpret feature maps of a VGG network, about 80\%–96\% activation scores came from interpretable features.}
\label{tab:p}
\vspace{-12pt}
\end{table}

\section{Conclusion and discussions}

In this paper, we have proposed a theoretical solution to a new explanation strategy, \emph{i.e.} learning an explainer network to disentangle and explain feature maps of a pre-trained performer network. Learning an explainer besides the performer does not decrease the discrimination power of the performer, which ensures the broad applicability. We have developed a simple yet effective method to learn the explainer, which guarantees the high interpretability of feature maps without using annotations of object parts or textures for supervision.

We divide the encoder of the explainer into an interpretable track and an ordinary track to reduce the risk of over-interpreting textures or noises as parts. Fortunately, experiments have shown that about 90\% of signals in the performer can be explained as parts.

{\small
\bibliographystyle{aaai}
\bibliography{TheBib}
}

\end{document}